\documentclass[11pt,a4paper]{article}
\usepackage[hyperref]{acl2021}
\usepackage{times}
\usepackage{latexsym}

\usepackage{microtype}
\usepackage{booktabs}
\usepackage{multirow}
\usepackage{amsmath}
\usepackage{bm}
\usepackage{amsfonts}
\usepackage{array}
\usepackage{graphicx}
\usepackage{color}

\aclfinalcopy 
\def\aclpaperid{65} 

\title{{F}ew-{S}hot {E}vent {D}etection with {P}rototypical {A}mortized \\ {C}onditional {R}andom {F}ield}

\author{Xin Cong$^{1,2}$ \ Shiyao Cui$^{1,2}$ \ Bowen Yu$^{1,2}$ \\ \bf{Tingwen Liu}$^{1,2}$\thanks{\hspace{0.15cm}Corresponding   Author} \ \bf{Yubin Wang}$^{1,2}$ \and \bf{Bin Wang}$^{3}$  \\
	$^1$Institute of Information Engineering, Chinese Academy of Sciences. Beijing, China \\
	$^2$School of Cyber Security, University of Chinese Academy of Sciences. Beijing, China \\
	$^3$Xiaomi AI Lab, Xiaomi Inc., Beijing, China \\
	{\tt \{congxin, cuishiyao, yubowen\}@iie.ac.cn} \\
	{\tt \{liutingwen, wangyubin\}@iie.ac.cn} \\
	{\tt wangbin11@xiaomi.com}
}

\date{}

\begin{document}
\maketitle

\begin{abstract}
Event detection tends to struggle when it needs to recognize novel event types with a few samples.
The previous work attempts to solve this problem in the identify-then-classify manner but ignores the trigger discrepancy between event types, thus suffering from the error propagation.
In this paper, we present a novel unified model which converts the task to a few-shot tagging problem with a double-part tagging scheme.
To this end, we first propose the \textbf{P}rototypical \textbf{A}mortized \textbf{C}onditional \textbf{R}andom \textbf{F}ield (PA-CRF) to model the label dependency in the few-shot scenario, which approximates the transition scores between labels based on the label prototypes.
Then Gaussian distribution is introduced for modeling of the transition scores to alleviate the uncertain estimation resulting from insufficient data.
Experimental results show that the unified models work better than existing identify-then-classify models and our PA-CRF further achieves the best results on the benchmark dataset FewEvent.
Our code and data are available at \url{http://github.com/congxin95/PA-CRF}.
\end{abstract}

\section{Introduction}

%
%

Event detection (ED) systems extract events of specific types from the given text.
Traditionally, researchers use pipeline approaches~\cite{ahn-2006-stages} where a trigger identification (TI) system is used to identify event triggers in a sentence and then a trigger classifier (TC) is used to find the event types of extracted triggers.
Such a framework makes the task easy to conduct but ignores the interaction and correlation between the two subtasks, being susceptible to cascading errors.
In the last few years, several neural network-based models were proposed to jointly identify triggers and classify event types from a sentence~\cite{chen-etal-2015-event,nguyen-grishman-2015-event,DBLP:conf/aaai/NguyenG18,liu-etal-2018-jointly,yan-etal-2019-event,cui2020EEGCN,cui2021LHGAT}.
These models have achieved promising performance and proved the effectiveness of solving ED in the joint framework.
But they almost followed the supervised learning paradigm and depended on the large-scale human-annotated dataset, while new event types emerge every day and most of them suffer from the lack of sufficient annotated data.
In the case of insufficient resources, existing joint models cannot recognize the novel event types with only few samples, i.e., Few-Shot Event Detection (FSED).

\begin{figure}[!t]
	\centering
	\includegraphics[scale=0.43]{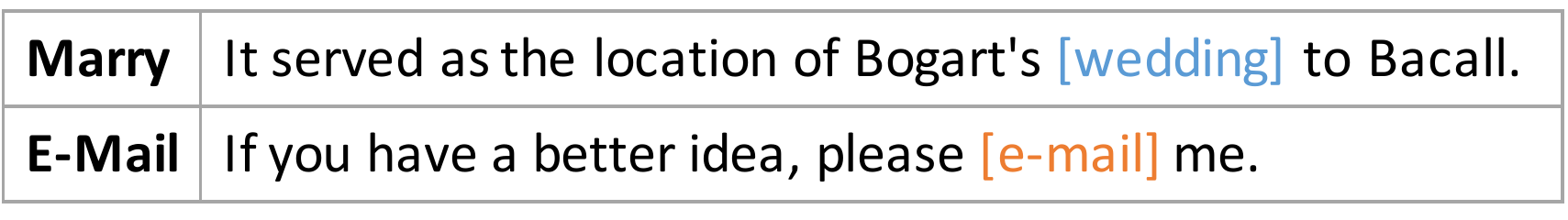}
	\caption{An example from FewEvent dataset revealing the trigger discrepancy. ``[$\cdot$]'' marks the event trigger.}
	\label{fig:example}
\end{figure}

One intuitive way to solve this problem is to first identify event triggers in the conventional way and then classify the event types based on the few-shot learning~\cite{vinyals2016matching,snell2017prototypical,DBLP:conf/cvpr/SungYZXTH18}, these two subtasks can be trained jointly by parameter sharing. 
Such identify-then-classify paradigm~\cite{DBLP:conf/wsdm/DengZKZZC20} seems to be convincing because TI aims to recognize triggers and does not need to adapt to novel classes, so we just need to solve the TC in the few-shot manner.
Unfortunately, our preliminary experiments reveal that TI tends to struggle when recognizing triggers of novel event types because novel events usually contain completely different triggers with the semantic distinction from the known events, i.e., \textbf{Trigger discrepancy} problem.
Figure \ref{fig:example} gives an example that the trigger ``e-mail'' would only occur in event \textit{E-Mail} but not in \textit{Marry} and triggers of two events have disparate context.
And experiments on FewEvent (a benchmark dataset for FSED) show that 59.21\% triggers in the test set do not trigger any events in the training set and the F1 score of TI with the SOTA TI model BERT-tagger~\cite{yang-etal-2019-exploring} is only 31.06\%.
Thus, the performance of the identify-then-classify paradigm will be limited by the TI part due to the cascading errors.

In this paper, we present a new unified method to solve FSED.
Specifically, we convert this task to a sequence labeling problem and design a double-part tagging scheme using trigger and event parts to describe the features of each word in a sentence.
The key to the sequence labeling framework is to model the dependency between labels.
Conditional Random Field (CRF) is a popular choice to capture such label dependency by learning transition scores of fixed label space in the training dataset.
Nevertheless, in FSED, CRF cannot be applied directly due to the \textbf{label discrepancy} problem, that is the label space of the test set is non-overlapping with the training set since FSED aims to recognize novel event types.
Therefore, the learned transition scores of CRF from the training set do not model the dependency of the novel labels in the test set.

%
%

To address the label discrepancy problem, we propose \textbf{P}rototypical \textbf{A}mortized \textbf{C}onditional \textbf{R}andom \textbf{F}ield (PA-CRF), which approximates the transition scores based on the label prototypes~\cite{snell2017prototypical} instead of learning by optimization.
Specifically, we first apply the self-attention mechanism to capture the dependency information between labels and then map the label prototype pairs to the corresponding transition scores.
In this way, PA-CRF can produce label-specific transition scores based on the few supportive samples, which can adapt to arbitrary novel event types.
However, predicting the transition score as a single fixed value actually acts as the point estimation, which usually acquires a large amount of annotated data to achieve accurate estimation.
Estimated from the handful of samples, the transition scores may suffer from the statistical uncertainty due to the random fluctuation of scant data.
To release this issue, 
inspired by variational inference~\cite{DBLP:journals/corr/KingmaW13,NEURIPS2018_e1021d43,gordon2018metalearning}, 
we treat the transition score as the random variable and utilize the Gaussian distribution to approximate its distribution to model the uncertainty.
Thus, our PA-CRF is to estimate the parameters of the Gaussian distribution rather than the transition scores directly, i.e., in the amortized manner~\cite{DBLP:journals/corr/KingmaW13,gordon2018metalearning}.
The Probabilistic Inference~\cite{gordon2018metalearning} is employed based on the Gaussian distribution to make the inference robust by taking the possible perturbation of transition scores into account since the perturbation is also learned in a way that coherently explains the uncertainty of the samples.

%
%

To summarize, our contributions are as follows: 
\begin{itemize}
	\item We devise a tagging-based unified model for FSED. To the best of our knowledge, we are the first to solve this task in a unified manner, free from the cascading errors.
	\item  We propose a novel model, PA-CRF, which estimates the distributions of transition scores for modeling the specific label dependency in the few-shot sequence labeling setting.
	\item Experimental results show that our proposed PA-CRF outperforms other competitive baselines on the FewEvent dataset. Further analyses show the effectiveness of our unified model and the limitation of the identify-then-classify models.
\end{itemize}

\section{Related Work}


%
%

\paragraph{Few-shot Event Detection} 
Event Detection (ED) aims to recognize the specific type of events in a sentence.
In recent years, various neural-based models have been proposed and achieved promising performance in ED~\cite{chen-etal-2015-event,nguyen-grishman-2015-event,DBLP:conf/aaai/NguyenG18,liu-etal-2018-jointly,yan-etal-2019-event,cui2020EEGCN}.
%
\citeauthor{chen-etal-2015-event}~\shortcite{chen-etal-2015-event} and \citeauthor{nguyen-grishman-2015-event}~\shortcite{nguyen-grishman-2015-event} proposed the convolution architecture to capture the semantic information in the sentence.
\citeauthor{nguyen-etal-2016-joint-event}~\shortcite{nguyen-etal-2016-joint-event} introduced the recurrent neural network to model the sequence contextual information of words.
Recently, GCN-based models~\cite{DBLP:conf/aaai/NguyenG18,liu-etal-2018-jointly,yan-etal-2019-event,cui2020EEGCN} have been proposed to exploit the syntactic dependency information and achieved state-of-the-art performance.
However, all these models are data-hungry, limiting dramatically their usability and deployability in real-world scenarios.

%
%
Recently, there has been an increasing research interest in solving event detection in the few-shot scenarios~\cite{DBLP:conf/wsdm/DengZKZZC20,DBLP:conf/pakdd/LaiDN20,DBLP:journals/corr/abs-2006-10093}, by exploiting the Few-Shot Learning ~\cite{vinyals2016matching,snell2017prototypical,DBLP:conf/icml/FinnAL17,DBLP:conf/cvpr/SungYZXTH18,cong2020DaFeC}.
\citeauthor{DBLP:conf/pakdd/LaiDN20}~\shortcite{DBLP:conf/pakdd/LaiDN20} proposed LoLoss which splits the part of the support set to act as the auxiliary query set to train the model.
\citeauthor{DBLP:journals/corr/abs-2006-10093}~\shortcite{DBLP:journals/corr/abs-2006-10093} introduced two regularization matching losses to improve the performance of models.
These works only focus on the few-shot trigger classification which classifies the event type of the annotated trigger according to the context based on few samples.
This is unrealistic as triggers of novel events are predicted by some existing toolkits in advance.
\citeauthor{DBLP:conf/wsdm/DengZKZZC20}~\shortcite{DBLP:conf/wsdm/DengZKZZC20} first proposed the benchmark dataset, \textit{FewEvent}, for FSED and designed the DMBPN based on the dynamic memory networks.
They train a conventional trigger identifier and a few-shot trigger classifier jointly and evaluated the model performance in the identify-then-classify paradigm.
Moreover, our preliminary experiments reveal that the conventional trigger identification model tends to struggle when recognizing triggers of novel event types because of the trigger discrepancy between different event types.
Thus, errors of the trigger identifier might be propagated to the event classification. 
Different from the previous identify-then-classify framework, for the first time, we solve Few-Shot Event Detection with two subtasks in a unified manner.

%
%

%
\paragraph{Few-shot Sequence Labeling} 
In recent years, several works~\cite{warmprotoz,hou-etal-2020-shot,DBLP:conf/emnlp/YangK20} have been proposed to solve the few-shot named entity recognition using sequence labeling methods.
\citeauthor{warmprotoz}~\shortcite{warmprotoz} applied the vanilla CRF in the few-shot scenario directly.
\citeauthor{hou-etal-2020-shot}~\shortcite{hou-etal-2020-shot} proposed a collapsed dependency transfer mechanism (CDT) into CRF, which learns label dependency patterns of a set of task-agnostic abstract labels and utilizes these patterns as transition scores for novel labels.
\citeauthor{DBLP:conf/emnlp/YangK20}~\shortcite{DBLP:conf/emnlp/YangK20} trains their model on the training data in a standard supervised learning manner and then uses the prototypical networks and the CDT for prediction in the inference phase.
Different from these methods learning the transition scores by optimization, we build a network to generate the transition scores based on the label prototypes instead.
In this way, we can generate exact label-specific transition scores of arbitrary novel event types to achieve adaptation ability.
And we further introduce the Gaussian distribution to estimate the data uncertainty.
Experiments prove the effectiveness of our method over the previous methods.

\section{Problem Formulation}

We convert event detection to a sequence labeling task. 
Each word is assigned a label that contributes to detecting the events. 
Labels consist of two parts: the word position in the trigger and the event type. 
We use the ``BI'' (Begin, Inside) signs to represent the position information of a word in the event trigger. 
The event type information is obtained from a predefined set of events. 
Label ``O'' (Other) means that the corresponding word is independent of the target events. 
Thus, the total number of labels is $2N+1$ ($N$ for \textit{B-EventType}, $N$ for \textit{I-EventType}, and an additional \textit{O} label), where $N$ is the number of predefined event types.

Furthermore, we formulate the Few-Shot Event Detection in the typical $N$-way-$K$-shot paradigm.
Let $\bm{x} = \{w_1, w_2, \ldots, w_n\}$ denote an $n$-word sequence, and $\bm{y} = \{ y_1, y_2, \ldots, y_n\}$ denote the label sequence of the $\bm{x}$.
Given a \textit{support set} $\mathcal{S} = \{ (\bm{x}^{(i)}, \bm{y}^{(i)}) \}_{i=1}^{N \times K}$ which contains $N$ event types and each event type has only $K$ instances, FSED aims to predict the labels of a unlabeled \textit{query set} $\mathcal{Q}$ based on the \textit{support set} $\mathcal{S}$.
Formally, a $\{ \mathcal{S}, \mathcal{Q} \}$ pair is called a $N$-way-$K$-shot task $\mathcal{T}$.
There exist two datasets consisting of a set of tasks : $\mathcal{D}_{train} = \{ \mathcal{T}^{(i)} \}_{i=1}^{M_{train}}$ and $\mathcal{D}_{test} = \{ \mathcal{T}^{(i)} \}_{i=1}^{M_{test}}$ where $M_{train}$ and $M_{test}$ denote the number of the task in two datasets respectively.
As the name suggests, $\mathcal{D}_{train}$ is used to train models in the training phase while $\mathcal{D}_{test}$ is for evaluation.
It is noted that these two datasets have their own event types, which means that the label space of two datasets is disjoint with each other.

\section{Methodology}

\begin{figure*}[!t]
	\centering
	\includegraphics[width=1\linewidth]{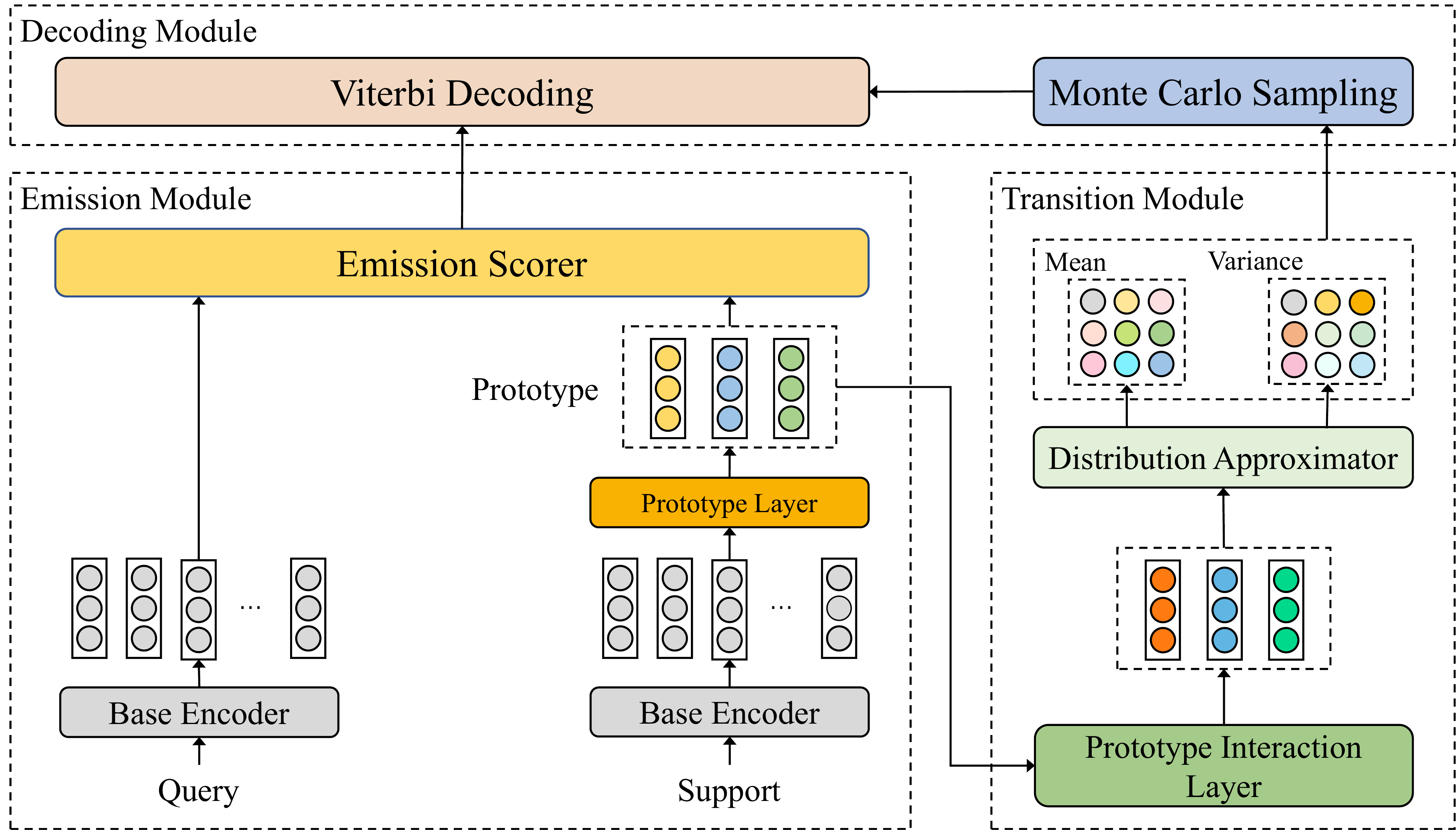}
	\caption{Architecture of our proposed PA-CRF. It consists of three modules: a) Emission Module calculates the emission scores for the query instance based on the prototypes derived from the support set. b) Transition Module generates the Gaussian distributed transition scores with respect to prototypes. c) Decoding Module exploits the emission scores and approximated Gaussian distributed transition scores to decode the predicted label sequence with the Monte Carlo Sampling.}
	\label{fig:model}
\end{figure*}

\subsection{Overview}

As described above, we formulate FSED as the few-shot sequence labeling task with interdependent labels.
Following the widely used CRF framework, we propose a novel PA-CRF model to model such label dependency in the few-shot setting, and decode the best-predicted label sequence.
Our PA-CRF contains three modules:
1) Emission Module: It first computes the prototype of each label based on the support set, and then calculates the similarity between prototypes and each token in the query set as the emission scores.
2) Transition Module: It exploits the prototypes to generate the parameters of Gaussian distribution of the transition scores for decoding.
3) Decoding Module: Based on the emission scores and Gaussian distributed transition scores, the Decoding Module calculates the probabilities of possible label sequences for the given query set and decodes the predicted label sequence.
Figure \ref{fig:model} gives an illustration of PA-CRF.
We detail each component from the bottom to the top.

\subsection{Emission Module}

The Emission Module assigns the emission scores to each token of sentences in the query set $\mathcal{Q}$ with regard to each label based on the support set $\mathcal{S}$.

%
%

\subsubsection{Base Encoder}

Base Encoder aims to embed tokens in both support set $\mathcal{S}$ and query set $\mathcal{Q}$ into real-value embedding vectors to capture the semantic information.

Since BERT~\cite{DBLP:conf/naacl/DevlinCLT19} shows its advanced ability to capture the sequence information and has been widely used in NLP tasks recently, we use it as the backbone.
Given an input word sequence $\bm{x}$,  BERT first maps all tokens into hidden embedding representations.
We denote this operation as:
\begin{equation}
	\small
	\{ \mathbf{h}_1, \mathbf{h}_2, \ldots, \mathbf{h}_n \} = {\rm BERT}(\bm{x})
\end{equation}
where $\mathbf{h}_i \in \mathbb{R}^{d_h}$ refers to the hidden representation of token $w_i$, $d_h$ is the dimension of the hidden representation.

%
%

\subsubsection{Prototype Layer}

Prototype Layer is to derive the prototypes of each label from the support set $\mathcal{S}$.
As described in the problem formulation, we use the BIO schema to annotate the event trigger and $N$ event types could contain $2N+1$ labels.
Thus, indeed, we could get $2N+1$ prototypes.
Following the previous work~\cite{snell2017prototypical}, we calculate the prototype of each label by averaging all the word representations with that label in the support set $\mathcal{S}$:
\begin{equation}
	\small
	\mathbf{c}_i = {\frac{1}{|\mathcal{S}(y_i)|} \sum_{w \in \mathcal{S}(y_i)}} \mathbf{h}, \quad i = 1, 2, \ldots, 2N+1, 
\end{equation}
where $\mathbf{c}_i$ denotes the prototype for label $y_i$, $\mathcal{S}(y_i)$ refers to the token set containing all words in the support set $\mathcal{S}$ with label $y_i$, $\mathbf{h}$ represents the corresponding hidden representation of token $w$, and $|\cdot|$ is the number of set elements.

%
%

\subsubsection{Emission Scorer}

Emission Scorer aims to calculate the emission score for each token in the query set $\mathcal{Q}$. 
The emission scores are calculated according to the similarities between tokens and prototypes.
The computation of the emission score of the label $y_i$ for the word $w_j$ is defined as:
\begin{equation}
	\small
	f_{E}(y_i, w_j, \mathcal{S}) = d( \mathbf{c}_i, \mathbf{h}_j),
\end{equation}
where $d(\cdot, \cdot)$ is the similarity function.
In practice, we choose the dot product operation to measure the similarity.

Finally, given a word sequence $\bm{x}$, the emission score of the whole sentence with its corresponding ground-truth label sequence $\bm{y}$ is computed as:
\begin{equation}
	\small
	{\rm EMIT}(\bm{y}, \bm{x}, \mathcal{S}) = \sum_{i=1}^n f_{E}(y_i, w_i, \mathcal{S}).
	\label{eqn:emit}
\end{equation}

\subsection{Transition Module}

In vanilla CRF, the transition scores are learnable parameters and optimized from large-scale data to model the label dependency.
However, in the few-shot scenarios, the learned transition scores cannot adapt to the novel label set due to the disjoint label space.
To overcome this problem, we use neural networks to generate the transition scores based on the label prototypes instead of learning transition scores by optimization to achieve adaptation ability.
In this case, a problem needing to be solved is that using few support instances with random data fluctuation to generate transition scores would cause uncertain estimation and result in wrong inference.
To model the uncertainty, we treat the transition score as a random variable and use the Gaussian distribution to approximate its distribution.
Specifically, the Transition Module is to generate the distributional parameters (mean and variance) of transition scores based on the label prototypes.
It consists of two layers: 1) Prototypical Interaction Layer and 2) Distribution Approximator.
Details of each layer are listed in the following parts.

%
%

\subsubsection{Prototype Interaction Layer}

Since the transition score is to model the dependency between labels, the individual prototype for each event type with rare dependency information is hard to generate their transition scores.
Thus, we propose a Prototype Interaction Layer which exploits the self-attention mechanism to capture the dependency between labels.

We first calculate the attention scores of each prototype $\bm{c}_i$ with others:
\begin{equation}
	\small
	\alpha_{ij} = \frac{{\rm exp}(\bm{c}^{(q)}_i \cdot \bm{c}^{(k)}_j)}{\sum_{m=1}^{2N+1} {\rm exp}(\bm{c}^{(q)}_i \cdot \bm{c}^{(k)}_m)},
\end{equation}
where $\bm{c}^{(q)}_i$ and $\bm{c}^{(k)}_i$ are transformed from $\bm{c}_i$ by two linear layers respectively:
\begin{equation}
	\small
	\begin{aligned}
		c^{(q)}_i &= W^{(q)} c_i + b^{(q)} \\
		c^{(k)}_i &= W^{(k)} c_i + b^{(k)}
	\end{aligned}
\end{equation}

Getting the attention scores, the prototype $\tilde{\bm{c}}_i$ with dependency information is calculated as follows:
\begin{equation}
	\small
	\tilde{\bm{c}}_i = \sum_{j=1}^{2N+1} \alpha_{ij} \bm{c}^{(v)}_j,
\end{equation}
where $\bm{c}^{(v)}_i$ is also transformed linearly from $\bm{c}_i$:
\begin{equation}
	\small
	c^{(v)}_i = W^{(v)} c_i + b^{(v)}
\end{equation}

\subsubsection{Distribution Approximator}

This module aims to generate the mean and variance of Gaussian distributions based on the prototypes with dependency information.

We first denote the transition score matrix as $T_r \in \mathbb{R}^{(2N+1) \times (2N+1)}$ for all label pairs, and denote the the $i$-th row $j$-th column element of $T_r$ as $[T_r]_{ij}$ which refers to the transition score for label $y_i$ transiting to label $y_j$.
As treating $[T_r]_{ij}$ as random variable, we use the Gaussian distribution $[\tilde{T_r}]_{ij} \sim \mathcal{N}(\mu_{ij}, \sigma^2_{ij})$ to approximate $[T_r]_{ij}$, where $\mathcal{N} (\cdot, \cdot)$ refers to the Gaussian distribution.
To estimate the mean $\mu_{ij}$ and variance $\sigma_{ij}$ of $[\tilde{T_r}]_{ij}$, we concatenate the corresponding prototypes $\tilde{c}_{i}$ and $\tilde{c}_{j}$ and feed into two feed-forward neural networks respectively:

\small
\begin{align}
	\mu_{ij} &= W^{(\mu)}[\tilde{\bm{c}}_i \| \tilde{\bm{c}}_j] + b^{(\mu)} \\
	\sigma^2_{ij} &= {\rm exp} \left( W^{(\sigma^2)}[\tilde{\bm{c}}_i \| \tilde{\bm{c}}_j] + b^{(\sigma^2)} \right)
\end{align}
\normalsize
where $[\cdot \| \cdot]$ means the concatenation operation.
Given a label sequence $\bm{y}$, the transition score of the whole label sequence is approximated by:
\begin{equation}
	\small
	{\rm TRANS}(\bm{y}, \tilde{T_r}) = \sum_{i=1}^{n-1} [\tilde{T_r}]_{\mathbb{I}(y_i)\mathbb{I}(y_{i+1})}
	\label{eqn:trans}
\end{equation}
where $\mathbb{I}(y_i)$ refers to the label index of $y_i$.

\subsection{Decoding Module}

Decoding Module derives the probabilities for a specific label sequence of the query set according to the emission scores and approximated Gaussian distributions of transition scores.

Since the approximated transition score is Gaussian distributional and not a single value, we denote the probability density function of the approximated transition score matrix as $q(\tilde{T_r} | \mathcal{S})$.
According to the Probabilistic Inference~\cite{gordon2018metalearning}, the probability of label sequence $\bm{y}$ of a word sequence $\bm{x}$ based on the support set $\mathcal{S}$ is calculated as:
\begin{equation}
	\small
	\begin{split}
		P(\bm{y}|\bm{x}, \mathcal{S}) &= \int P(\bm{y}|\bm{x}, \mathcal{S}, \tilde{T_r}) q(\tilde{T_r} | \mathcal{S}) \mathrm{d}\tilde{T_r} 
	\end{split}
\end{equation}

Following the CRF algorithm, the probability can be calculated based on the Equation \ref{eqn:emit} and Equation \ref{eqn:trans}:
\begin{equation}
	\small
	\begin{split}
		& P(\bm{y}|\bm{x}, \mathcal{S})  = \\
		& \int \frac{1}{Z} {\rm exp} \left( {\rm EMIT}(\bm{y}, \bm{x}, \mathcal{S}) + {\rm TRANS}(\bm{y}, \tilde{T_r}) \right) q(\tilde{T_r} | \mathcal{S}) \mathrm{d}\tilde{T_r}
	\end{split}
	\label{eqn:intergral}
\end{equation}
where 
\begin{equation}
	\small
	Z = \sum_{\bm{y'} \in Y} {\rm exp} \left( {\rm EMIT}(\bm{y'}, \bm{x}, \mathcal{S}) + {\rm TRANS}(\bm{y'}, \tilde{T_r}) \right)
\end{equation}
and $Y$ refers to all possible label sequences.

In the training phase, we use negative log-likelihood loss as our objective function:
\begin{equation}
	\small
	\mathcal{L} = - \mathop{\mathbb{E}}\limits_{(\bm{x}, \bm{y}) \sim \mathcal{Q}} \left[ {\rm log}(P(\bm{y}|\bm{x}, \mathcal{S})) \right]
\end{equation}
Due to the hardness to compute the integral of Equation \ref{eqn:intergral}, in practice, we use the Monte Carlo sampling technique~\cite{gordon2018metalearning} to approximate the integral.
To make the sampling process differentiable for optimization, we employ the reparameterization trick~\cite{DBLP:journals/corr/KingmaW13} for each transition score $[\tilde{T_r}]_{ij}$:
\begin{equation}
	\small
	[\tilde{T_r}]_{ij} = \mu_{ij} + \epsilon \sigma_{ij}, {\rm where} \ \epsilon \sim \mathcal{N}(0, 1)
\end{equation}

In the inference phase, the Viterbi algorithm~\cite{viterbi} is employed to decode the best-predicted label sequence for the query set.

\section{Experiment}

\subsection{Dataset}

We conduct experiments on the benchmark \textit{FewEvent} dataset introduced in the previous work~\cite{DBLP:conf/wsdm/DengZKZZC20}, which is the currently largest few-shot dataset for event detection.
It contains 70,852 instances for 100 event types and each event type owns about 700 instances on average.
Since \citeauthor{DBLP:conf/wsdm/DengZKZZC20}~\shortcite{DBLP:conf/wsdm/DengZKZZC20} do not share their split train/dev/test set, we re-split the FewEvent in the same ratio as \citeauthor{DBLP:conf/wsdm/DengZKZZC20}~\shortcite{DBLP:conf/wsdm/DengZKZZC20}.
%
We use 80 event types as the training set, 10 event types as the dev set, and the rest 10 event types as the test set.
More statistics of FewEvent dataset are listed in Appendix~\ref{sec:dataset}.

\subsection{Evaluation}

We follow the evaluation metrics in previous event detection works~\cite{chen-etal-2015-event,liu-etal-2018-jointly,cui2020EEGCN}, an event trigger is marked correct if and only if its event type and its offsets in the sentence are both correct.
We adopt the standard micro F1 score to evaluate the results and report the averages and standard deviations over 5 randomly initialized runs.

\section{Implementation Details}

We employ \textit{BERT-BASE-UNCASED}~\cite{DBLP:conf/naacl/DevlinCLT19} as the base encoder.
The maximum sentence length is set as 128.
Our model is trained using \textit{AdamW} optimizer with the learning rate of 1e-5.
All the hyper-parameters are tuned on the dev set manually.
In the training phase, we follow the widely used episodic training~\cite{vinyals2016matching} in few-shot learning. Episodic training aims to mimic N-way-K-shot scenario in the training phase. In each epoch, we randomly sample N event types from the training set and each event type randomly sample K instances as support set and other M instances as the query set.
We train our model with 20,000 iterations on the training set and evaluate its performance with 3,000 iterations on the test set following the episodic paradigm.
We run all experiments using PyTorch 1.5.1 on the Nvidia Tesla T4 GPU, Intel(R) Xeon(R) Silver 4110 CPU with 256GB memory on Red Hat 4.8.3 OS.
%

\subsection{Baselines}

\begin{table*}[!tb]
	\centering
	\begin{tabular}{c|l|cccc}  
		\toprule
		Paradigm & Model  & 5-Way-5-Shot & 5-Way-10-Shot & 10-Way-5-Shot & 10-Way-10-Shot\\
		\midrule
		Fine-tuning & PLMEE & 4.43 $\pm$ 0.19 & 4.69 $\pm$ 0.85 & 2.52 $\pm$ 0.28 & 2.76 $\pm$ 0.55 \\
		\midrule
		\multirow{2}*{Separate} & LoLoss		& 30.14 $\pm$ 0.30 & 30.91 $\pm$ 0.29 & 29.33 $\pm$ 0.40 & 30.08 $\pm$ 0.39 \\
		& MatchLoss	& 29.78 $\pm$ 0.14 & 30.75 $\pm$ 0.15 & 28.75 $\pm$ 0.23 & 29.59 $\pm$ 0.21 \\
		\midrule
		\multirow{3}*{Multi-task} & LoLoss		& 31.51 $\pm$ 1.56 & 31.70 $\pm$ 1.21 & 30.46 $\pm$ 1.38 & 30.32 $\pm$ 0.89 \\
		& MatchLoss	& 30.44 $\pm$ 0.99 & 30.68 $\pm$ 0.78 & 28.97 $\pm$ 0.61 & 30.05 $\pm$ 0.93 \\
		& DMBPN		& 37.51 $\pm$ 2.60 & 38.14 $\pm$ 2.32 & 34.21 $\pm$ 1.45 & 35.31 $\pm$ 1.69 \\
		\midrule
		\multirow{8}*{Unified} & Match 		& 39.93 $\pm$ 1.67 & 46.02 $\pm$ 1.20 & 30.88 $\pm$ 1.08 & 35.91 $\pm$ 1.19 \\
		& Proto 	& 50.11 $\pm$ 0.77 & 52.97 $\pm$ 0.95 & 43.51 $\pm$ 1.16 & 42.70 $\pm$ 0.98 \\
		& Proto-Dot & 58.82 $\pm$ 0.88 & 61.01 $\pm$ 0.23 & 55.04 $\pm$ 1.62 & 58.78 $\pm$ 0.88 \\
		& Relation 	& 28.91 $\pm$ 1.13 & 29.83 $\pm$ 0.78 & 18.49 $\pm$ 1.25 & 21.47 $\pm$ 1.40 \\
		\cmidrule{2-6}
		& Vanilla CRF	& 59.01 $\pm$ 0.81 & 62.21 $\pm$ 1.94 & 56.00 $\pm$ 1.51 & 59.35 $\pm$ 1.09 \\
		& CDT	& \underline{59.30} $\pm$ 0.23 & \underline{62.77} $\pm$ 0.12 & \underline{56.41} $\pm$ 1.09 & \underline{59.44} $\pm$ 1.83 \\
		& StructShot & 57.69 $\pm$ 0.91 & 61.54 $\pm$ 1.23 & 54.54 $\pm$ 0.95 & 57.14 $\pm$ 0.79 \\
		\cmidrule{2-6}
		& PA-CRF	& \textbf{62.25}* $\pm$ 1.42 & \textbf{64.45}* $\pm$ 0.49 & \textbf{58.48}* $\pm$ 0.68 & \textbf{61.64}* $\pm$ 0.81\\
		\bottomrule
	\end{tabular}
	\caption{F1 scores ($10^{-2}$) of different models on the FewEvent test set. Bold marks the highest number among all models, underline marks the second-highest number, and $\pm$ marks the standard deviation. * marks statistically significant improvements over the best baseline with $p < 0.01$ under a boostrap test.}	 
	\label{tab:main}
\end{table*}

To investigate the effectiveness of our proposed method, we compare it with a range of baselines and state-of-the-art models, which can be categorized into three classes: fine-tuning paradigm, identify-then-classify paradigm and unified paradigm.

%
%

\textbf{Fine-tuning paradigm} solves the FSED in the standard supervised learning, i.e., pre-training on the large scale dataset and fine-tuning on the handful target data.
We adopt the state-of-the-art model, \textbf{PLMEE}~\cite{yang-etal-2019-exploring}, of the standard ED into the FSED directly.

%
%

\textbf{Identify-then-classify models} first perform trigger identification (named as TI) and then classify the event types based on the few-shot learning methods (named as FSTC). 
We investigate two typed of identify-then-classify paradigms: separate and multi-task.
For the separate models, the trigger identifier and few-shot trigger classifier are trained separately without parameter sharing.
We first exploit the state-of-the-art BERT-tagger for the TI task.
It uses BERT~\cite{DBLP:conf/naacl/DevlinCLT19} and a linear layer to tag the trigger in the sentence as a sequence labeling task.
Since TI just aims to recognize the occurrence of the trigger, the label set only contains three labels: \textit{O}, \textit{B-Trigger}, \textit{I-Trigger}.
For the FSTC task, we reimplement two FSTC models: \textbf{LoLoss}~\cite{DBLP:conf/pakdd/LaiDN20}, \textbf{MatchLoss}~\cite{DBLP:journals/corr/abs-2006-10093}.
In the multi-task models, we reimplement \textbf{DMBPN}~\cite{DBLP:conf/wsdm/DengZKZZC20} and replace its encoder with BERT for the fair comparison.
DMBPN combines a conventional trigger identification module and a few-shot trigger classification module by parameter sharing.
But in the inference phase, it detects the event trigger still in the identify-then-classify paradigm.
Additionally, we also provide the multi-task version of the LoLoss and MatchLoss which are trained jointly with BERT-tagger with shared BERT parameters.

%
%

\textbf{Unified models} perform few-shot event detection with a single model without task decomposition.
Because we are the first to solve this task in a unified way, there is no previous unified model that can be compared.
But for the comprehensive evaluation of our proposed PA-CRF model, we also construct two groups of variants of PA-CRF: non-CRF models and CRF-based models.
Non-CRF models use emission scores to predict via softmax and do not take the label dependency into consideration.
We implement four typical few-shot classifiers:
1) \textbf{Match}~\cite{vinyals2016matching} uses cosine function to measure the similarity,
2) \textbf{Proto}~\cite{snell2017prototypical} uses Euclidean Distance as the similarity metric,
3) \textbf{Proto-Dot} uses dot product to compute the similarity,
4) \textbf{Relation}~\cite{DBLP:conf/cvpr/SungYZXTH18} builds a two-layer neural networks to measure the similarity.
%
%
Since CRF with the capacity of modeling label dependency is widely used in sequence labeling task, we implement three kinds of CRF-based models as our baselines: 
1) \textbf{Vanilla CRF}~\cite{warmprotoz}: We adopt the vanilla CRF in the FSED task without considering the adaptation problem. 
2) \textbf{CDT}~\cite{hou-etal-2020-shot}: As the SOTA of the few-shot NER task,
we re-implement it according to the official code and adapt it in the FSED task to replace our Transition Module.
3) \textbf{StructShot}~\cite{DBLP:conf/emnlp/YangK20}: It is also a few-shot NER model. It first pre-trains on the training set and utilizes the prototypical networks and the CDT for prediction based on the support set in the inference phase.
For the fair comparison, the emission module of these CRF-based baseline models is the same as our PA-CRF.

\subsection{Main Results}

Table \ref{tab:main} summarizes the results of our PA-CRF against other baseline models on the FewEvent test set.

\textbf{Comparison with fine-tuning model}
It is obvious that PLMEE performs poorly in all four few-shot settings and all few-shot-based methods outperform it with an absolute gap, which powerfully proves that the conventional supervised methods is incapable of solving FSED.

\textbf{Comparison with identify-then-classify models}
(1) Most of unified models (except Relation) perform higher than all identify-then-classify models, especially for PA-CRF with huge gaps about 30\%, proving the effectiveness of the unified framework.
(2) Comparing with the separate paradigm, the multi-task paradigm is able to improve performance but it still cannot catch up with the unified paradigm.
(3) DMBPN works better than other two models but still works poorly to handle the FSED due to the limitation of the TI.
%
We will discuss the bottleneck of the identify-then-classify paradigm in Section \ref{sec:bottleneck}.

%
%

\textbf{Comparison with unified models}
(1) Over the best non-CRF baseline model Proto-Dot, PA-CRF achieves substantial improvements of 3.43\%, 3.44\%, 3.44\% and 2.86\% on four few-shot scenarios respectively, which confirms the effectiveness and rationality of PA-CRF to model the label dependency.
(2) Vanilla CRF performs better than other non-CRF baseline methods, which demonstrates that CRF is able to improve the performance by modeling the label dependency, even if the learned transition scores do not match the label space of the test set.
%
(3) Compared to Vanilla CRF, both CDT and StructShot achieve slightly higher F1 scores, indicating the transition scores of abstract BIO labels can improve the model adaptation ability to some extent.
(4) CDT exceeds the StructShot since CDT is trained based on the episodic training, which makes it learns the class-agnostic token representations.
(5) PA-CRF outperforms CDT (2.95\%, 1.68\%, 2.07\% and 2.20\% in four few-shot settings respectively) with absolute gaps.
We consider that it is because CDT learning the transition scores of the abstract labels cannot model the exact dependency of specific label set, so its adaptation ability is limited.
In contrast, PA-CRF generates the label-specific transition scores based on the label prototype, which can capture the dependency for specific novel event types.
(6) Comparing four few-shot scenarios, we can find that the F1 score increases as the K-shot increases, which shows that more support samples can provide more information of the event type.
The F1 score decreases as the N-way increases when the shot number is fixed, which reveals that the larger way number causes more event types to predict which increases the difficulty of the correct detection.

%
%

To summarize, we can draw the conclusion that 
(1) The identify-then-classify paradigm is incapable of solving the FSED task.
(2) Compared to the identify-then-classify paradigm, the unified paradigm works more effectively for the FSED task.
(3) Approximating transition scores based on the label prototypes not by optimization, our PA-CRF achieves better adaptation on novel event types.

\subsection{Bottleneck Analysis}

\label{sec:bottleneck}

\begin{table}[!t]
	\centering
	\begin{tabular}{lccc}  
		\toprule
		Model  & TI & FSTC & FSED \\
		\midrule
		LoLoss				& 31.06 & 95.27 & 30.14 \\
		DMBPN				& 40.64	& 95.44 & 37.51 \\
		DMBPN(CDT-TI)		& 54.69 & 95.49 & 53.93 \\
		PA-CRF				& 63.68 & 96.76 & 62.25 \\
		\bottomrule
	\end{tabular}
	\caption{Comparison of PA-CRF and baselines on two subtasks. F1 scores are reported on the FewEvent test set in the 5-way-5-shot setting.}
	\label{tab:bottleneck}
\end{table}

To investigate the bottleneck of the identify-then-classify paradigm, we evaluate LoLoss (separate model), DMBPN (multi-task model) and PA-CRF (unified model) on two subtasks: TI and FSTC separately in the 5-way-5-shot setting on the FewEvent test set.
To reduce the influence of the cascading errors, we use the ground truth trigger span for evaluation in the FSTC.
The experimental results are reported in Table \ref{tab:bottleneck}.
From Table \ref{tab:bottleneck}, we find that:
(1) All models achieve more than 95\% F1 score on the FSTC task, indicating that both identify-then-classify and unified models is capable enough of solving the FSTC problem.
(2) For the TI task, two identify-then-classify baselines perform 31.06\% and 40.64\% F1 score respectively, which demonstrates that the conventional TI module has difficulty in adapting to novel event triggers.
Hence, due to the cascading errors, the poorly-performed TI module limits the performance of the identify-then-classify models.
(3) PA-CRF achieves 63.68\% F1 score on TI task, which exceeds the two kinds of identify-then-classify models significantly.
Unlike identify-then-classify models recognizing triggers based on seen triggers, PA-CRF utilizes the trigger representations from the support set of the novel event types to identify novel triggers so our unified model works better in the TI task of FSED.
In conclusion, the conventional trigger identifier cannot identify novel triggers in FSED, and exploiting the support set of novel event types is necessary.

%
%

\subsection{Effectiveness Analysis}

To verify the effectiveness of the unified framework, we adapt our best baseline model, CDT, to replace TI module of DMBPN to solve trigger identification in the few-shot manner.
It identifies triggers based on the emission scores between tokens and label prototypes calculating from the support set and learned abstract transition scores.
In this case, we rename it as DMBPN(CDT-TI) and evaluate it in TI and FSTC subtasks.
Results are also reported based on the 5-way-5-shot setting in Table \ref{tab:bottleneck} and we observe that:
The CDT-TI-based DMBPN achieve 54.69\% in TI task, exceeding the conventional TI based models, which shows that solving TI in the few-shot manner by utilizing the support set can reduce the trigger discrepancy to some extent.
Although the performance of FSTC is similar to the original DMBPN, owing to the improvements of TI task, the final performance of FSED exceeds the original DMBPN by 16.42\% but they are still inferior to PA-CRF with a huge gap (8.99\% on TI task).
Therefore, we draw the conclusion that solving FSED in the unified manner can utilize the correlation between two subtasks to improve the model performance significantly.

\subsection{Ablation Study}

\begin{table}[!tb]
	\centering
	\begin{tabular}{lcc}  
		\toprule
		Model 	& 5-Shot & 10-Shot \\
		\midrule
		PA-CRF								& 44.39 & 51.06 \\
		\enspace - Distribution Estimation	& 43.47 & 49.41 \\
		\enspace - Interaction Layer		& 41.62 & 45.74 \\
		\enspace - Transition Score			& 39.83 & 45.07 \\
		\bottomrule
	\end{tabular}
	\caption{Ablation study of PA-CRF in 5-Way settings. F1 scores are reported on the FewEvent dev set.}
	\label{tab:ablation}
\end{table}

To study the contribution of each component in our PA-CRF model, we run the ablation study on the FewEvent dev set. 
From these ablations (see Table~\ref{tab:ablation}), we find that:
(1) - Distribution Estimation: To study whether distributional estimation is helpful to improve the performance, we remove it and make the Distribution Approximator generate a single value as the transition score directly as the point estimation.
And the inference is based on the generated transition scores without Probabilistic Inference.
As a result, the F1 score drops 1.02\% and 1.65\% in two scenarios, respectively.
We attribute these gaps to our proposed Gaussian-based distributional estimation which can model the data fluctuation to relieve the influence of data uncertainty.
(2) - Interaction Layer: To certify that the Prototype Interaction Layer contributes to capturing the information between prototypes, we remove it and evaluate in two scenarios.
We read from Table~\ref{tab:ablation} that F1 scores decrease significantly by 2.77\% and 5.32\% respectively, which indicates that the Prototype Interaction Layer is able to capture the dependency among prototypes.
(3) - Transition Score: To prove the contribution of the label dependency, we remove the Transition Module and only use the emission score for prediction.
Results show that without transition scores, the performance of the model drops dramatically by 4.56\% and 5.99\% respectively, which powerfully proves that the transition score can improve the performance of the few-shot sequence labeling task.

Furthermore, we have conducted case study and error analysis to validate the strength of our PA-CRF and explore its weakness. Details are listed in Appendix~\ref{sec:case_study} and Appendix~\ref{sec:error_study}.

\section{Conclusion}

In this paper, we explore a new viewpoint of solving few-shot event detection in a unified manner.
Specifically, we propose a prototypical amortized conditional random field to generate the transition scores to achieve adaptation ability for novel event types based on the label prototypes.
Furthermore, we present the Gaussian-based distributional estimation to approximate transition scores to relieve the statistical uncertainty of data fluctuation.
Finally, experimental results on the benchmark FewEvent dataset prove the effectiveness of our proposed method. 
In the future, we plan to adapt our method to other few-shot sequence labeling tasks such as named entity recognition.

\section*{Acknowledgements}

We would like to thank all reviewers for their insightful comments and suggestions.
This work is supported in part by the Strategic Priority Research Program of Chinese Academy of Sciences (grant No. XDC02040400) and the Youth Innovation Promotion Association of CAS (Grant No. 2021153).

\bibliographystyle{acl_natbib}
\bibliography{anthology,ref}

\appendix

\section{Dataset Statistics}
\label{sec:dataset}

\begin{table}[!h]
	\centering
	\begin{tabular}{lccc}  
		\toprule
		& Training Set & Dev Set & Test Set \\
		\midrule
		\# Cls.	 		& 80 & 10 & 10 \\
		\# Sent.		& 67982 & 2173 & 697 \\
		\# Tok./sent	& 36.5 & 38.6 & 30.8 \\
		\bottomrule
	\end{tabular}
	\caption{Statistics of FewEvent Dataset.}
	\label{tab:stat}
\end{table}

Table \ref{tab:stat} lists the statistics of FewEvent dataset containing the number of event type (\#Cls.), the number of sentence(\# Sent.), the number of token per sentence (\# Tok./sent) for the train/dev/test set.

\begin{figure*}[!tbh]
	\centering
	\includegraphics[scale=0.35]{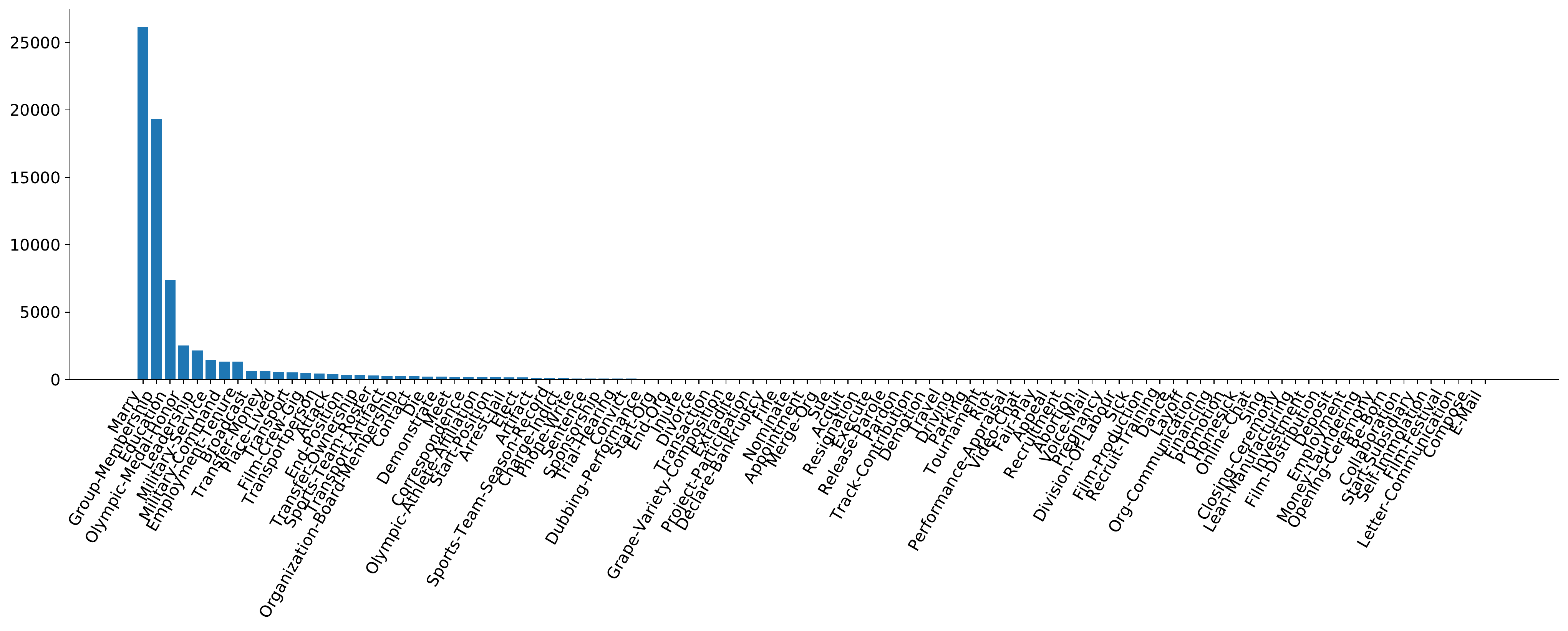}
	\caption{The data imbalance of FewEvent dataset.}
	\label{fig:stat}
\end{figure*}

Figure \ref{fig:stat} demonstrates the data imbalance problem of FewEvent dataset. 
Event ``Marry'' has the most instance (26135 instances) while event ``E-Mail'' only has 30 instances.
69\% event types have less than 100 instances while 7\% event types have more than 1000 instances.
However, since we use episodic training~\cite{vinyals2016matching} to train our model, the data imbalance problem can be relieved to some extent.

\section{Case Study}
\label{sec:case_study}

\begin{table*}
	\centering
	\begin{tabular}{l|l}
		\toprule
		Model & \multicolumn{1}{c}{Prediction} \\
		\midrule
		DBMPN & Candlestick Park was dropped when the \textcolor{gray}{[\textit{sponsorship}]$_{O}$} agreement expired. \\
		PA-CRF & Candlestick Park was dropped when the \textcolor{orange}{[\textit{sponsorship}]$_{B-Sponsorship}$} agreement expired. \\
		\midrule
		CDT & Willmore will tell everyone for wanting to keep the poor man \textcolor{blue}{[\textit{locked}]$_{I-Jail}$} \textcolor{blue}{[\textit{up}]$_{I-Jail}$}. \\
		PA-CRF & Willmore will tell everyone for wanting to keep the poor man \textcolor{orange}{[\textit{locked}]$_{B-Jail}$} \textcolor{blue}{[\textit{up}]$_{I-Jail}$}. \\
		\bottomrule
	\end{tabular}
	\caption{Output of PA-CRF, DMBPN and CDT on samples from the FewEvent test set. The subscripts denote the labels tagged by the models.}
	\label{tab:case}
\end{table*}

We compare our method with the best identify-then-classify baseline, DMBPN and the best unified baseline, CDT in some cases, as shown in Table \ref{tab:case}.

%
%

As demonstrated by the first example of a \textit{Sponsorship} event, DMBPN, in the identify-then-classify paradigm, fails to identify the trigger \textit{sponsorship}.
According to our statistics about the FewEvent dataset, 95.16\% triggers of \textit{Sponsorship} event do not occur in the training set.
%
%
Since DMBPN uses the conventional TI module which is trained on the training set to identify the event trigger, it is incapable of identifying the \textit{Sponsorship} event trigger. 
%
%
Although the classification module of DMBPN succeeds to distinguish the event type as \textit{Sponsorship}, due to the cascading errors, the final prediction of event trigger (containing the span and type) is incorrect.
As a result, the performance of DMBPN as an identify-then-classify model on the FSED task is limited. 
In contrast, our unified PA-CRF is successful to detect the event trigger \textit{sponsorship} of this case since PA-CRF utilizes the information of the support set of \textit{Sponsorship} event in which word \textit{sponsorship} appears and acts the trigger.

%
%

In the second example, the best unified baseline, CDT, tags the first trigger word \textit{locked} with \textit{I-Jail} label wrongly.
That is because CDT learns the abstract transition scores among a set of abstract labels which cannot model the label dependency for this specific event type accurately.
Thanks to the PA-CRF which models the label dependency based on the label prototypes from the support set of \textit{Jail} event, our model is capable of tagging the word \textit{locked} with \textit{B-Jail} label correctly. 

\section{Error Study}
\label{sec:error_study}

\begin{table*}[!t]
	\centering
	\begin{tabular}{l|l}
		\toprule
		Support \#1 & Cult members \textcolor{orange}{[\textit{visited}]$_{B-Trans}$} and built a laser weapon mounted on a truck \\
		Support \#2 & Israel \textcolor{orange}{[\textit{leave}]$_{B-Trans}$} the West Bank and Gaza and dismantle Jewish settlements. \\
		\midrule
		Truth & Refugees have been \textcolor{orange}{[\textit{pouring}]$_{B-Trans}$} \textcolor{blue}{[\textit{out}]$_{I-Trans}$} of Fallujah over the last few days. \\
		Prediction & Refugees have been \textcolor{orange}{[\textit{pouring}]$_{B-Trans}$} \textcolor{gray}{[\textit{out}]$_{O}$} of Fallujah over the last few days. \\
		\bottomrule
	\end{tabular}
	\caption{A case of the wrong prediction from the FewEvent test set. The subscripts denote the triggers and their event types. We only list two support instances to reduce space.}
	\label{tab:error}
\end{table*}

Although our method outperforms all baseline models, we still observe some failure cases.
Table \ref{tab:error} gives a typical example of the wrong prediction of event \textit{Transport} (\textit{Trans} for short).
For the query instance, the ground truth event trigger is ``pouring out''.
The word ``pouring'' should be labeled as \textit{B-Trans} and the \textit{out} should be labeled as \textit{I-Trans}.
However, our model only detects ``pouring'' with \textit{B-Trans} while missing ``out''.
From the support set, we find that all support instances of this event type only contain the one-word trigger without \textit{I-Trans} label tokens, resulting in that the prototype of \textit{I-Trans} is zero vector.
As a result, the emission score for the label \textit{I-Trans} of each query token is calculated as zero and the transition scores based on the prototypes are also affected.
Therefore, our model is not able to detect the \textit{I-Trans} label correctly in this case.
In the future, we will further study to solve the missing \textit{I} label problem.

\section{Analyses about Various Dataset Split}
\label{sec:dataset_split}

\begin{table}[!h]
	\centering
	\begin{tabular}{l|ccccc}
		\toprule
		Model & R1 & R2 & R3 & R4 & R5 \\
		\midrule
		PA-CRF & 59.0 & 33.4 & 53.1 & 42.4 & 48.0 \\
		DMBPN & 44.9 & 31.1 & 40.8 & 32.6 & 27.9 \\
		\bottomrule
	\end{tabular}
	\caption{Performance of our PA-CRF and DMBPN in various split FewEvent dataset in the 5-Way-5-Shot scenario. F1 scores ($10^{-2}$) are reported.}
	\label{tab:various_perform}
\end{table}

Since \citeauthor{DBLP:conf/wsdm/DengZKZZC20}~\shortcite{DBLP:conf/wsdm/DengZKZZC20} do not public their split train/dev/test set of FewEvent dataset, to compare our PA-CRF with DMBPN~\cite{DBLP:conf/wsdm/DengZKZZC20}, we re-split the FewEvent randomly in the same split ratio as the \citeauthor{DBLP:conf/wsdm/DengZKZZC20}~\shortcite{DBLP:conf/wsdm/DengZKZZC20} (80 event types for training set, 10 event types for dev set and the rest 10 event types for test set) and evaluate the DBMPN performance on our split test set.
However, in our experiments, the performance of DMBPN is lower than the original paper.
We assume that the different data split may influence the model performance badly.
To validate our assumption, we re-split the FewEvent dataset for five random seeds and conduct more experiments on these various split train/dev/test set.
The results are reported in Table \ref{tab:various_perform}.
From Table \ref{tab:various_perform}, it can be observed that:
(1) Data split does influence the model performance significantly indeed.
In these five different split train/dev/test set, the performance of PA-CRF varies from 59.0\% to 33.4\% with a huge range.
Similarly, DMBPN also varies from 44.9\% to 27.9\%, owning a huge gap about 20\%.
It demonstrates that for the FewEvent dataset, different split could cause huge fluctuation of the model performance.
Therefore, our PA-CRF including baselines performs lower than those \citeauthor{DBLP:conf/wsdm/DengZKZZC20}~\shortcite{DBLP:conf/wsdm/DengZKZZC20} reported due to the different data split.
(2) PA-CRF outperforms DMBPN in all five random split settings, which powerfully proves the robustness of PA-CRF over the identify-then-classify paradigm.

\end{document}